\documentclass[10pt,twocolumn,letterpaper]{article}

\usepackage[pagenumbers]{cvpr} 

\definecolor{cvprblue}{rgb}{0.21,0.49,0.74}
\usepackage[pagebackref,breaklinks,colorlinks,allcolors=cvprblue]{hyperref}

\usepackage{booktabs,tabularx,ragged2e,makecell,enumitem}

\usepackage{graphics} 
\usepackage{epsfig} 
\usepackage{mathptmx} 
\usepackage{times} 
\usepackage{amsmath} 
\usepackage{amssymb}  
\usepackage{siunitx}
\usepackage{booktabs}
\usepackage{multirow}
\usepackage{datatool}

\def\paperID{*****} 
\def\confName{CVPR}
\def\confYear{2026}

\title{Efficient Onboard Spacecraft Pose Estimation with Event Cameras and Neuromorphic Hardware}

\author{Arunkumar Rathinam$^{1}$, Jules Lecomte$^{2}$, Jost Reelsen$^{3}$, and Gregor Lenz$^{4}$%
\thanks{$^{1}$ University of Luxembourg, Esch-sur-Alzette, Luxembourg
        {\tt\small arunkumar.rathinam@uni.lu}}%
\thanks{$^{2}$ Fortiss GmbH, Munich, Germany
        {\tt\small juleslcmt@gmail.com}}%
\thanks{$^{3}$ Department of Natural Sciences, Technical University of Munich, Munich, Germany
        {\tt\small jost.reelsen@tum.de}}%
\thanks{$^{4}$ Paddington Robotics, London, United Kingdom
        {\tt\small lenz.gregor@gmail.com}}%
}
\author{Arunkumar Rathinam\\
University of Luxembourg\\
Luxembourg\\
{\tt\small arunkumar.rathinam@uni.lu}
\and
Jules Lecomte\\
Fortiss GmbH\\
Munich, Germany\\
{\tt\small juleslcmt@gmail.com}
\and
Jost Reelsen\\
Technical University of Munich\\
Munich, Germany\\
{\tt\small jost.reelsen@tum.de}
\and
Gregor Lenz\\
Paddington Robotics\\
London, United Kingdom\\
{\tt\small lenz.gregor@gmail.com}
\and
Axel von Arnim\\
Fortiss GmbH\\
Munich, Germany\\
{\tt\small vonarnim@fortiss.org}
\and
Djamila Aouada\\
University of Luxembourg\\
Luxembourg\\
{\tt\small djamila.aouada@uni.lu}
}
\begin{document}
\maketitle
\begin{abstract}
Reliable relative pose estimation is a key enabler for autonomous rendezvous and proximity operations, yet space imagery is notoriously challenging due to extreme illumination, high contrast, and fast target motion. Event cameras provide asynchronous, change-driven measurements that can remain informative when frame-based imagery saturates or blurs, while neuromorphic processors can exploit sparse activations for low-latency, energy-efficient inferences. This paper presents a spacecraft 6-DoF pose-estimation pipeline that couples event-based vision with the BrainChip Akida neuromorphic processor. Using the SPADES dataset, we train compact MobileNet-style keypoint regression networks on lightweight event-frame representations, apply quantization-aware training (8/4-bit), and convert the models to Akida-compatible spiking neural networks. We benchmark three event representations and demonstrate real-time, low-power inference on Akida V1 hardware. We additionally design a heatmap-based model targeting Akida V2 and evaluate it on Akida Cloud, yielding improved pose accuracy. To our knowledge, this is the first end-to-end demonstration of spacecraft pose estimation running on Akida hardware, highlighting a practical route to low-latency, low-power perception for future autonomous space missions.
\end{abstract}
\vspace{-1em}
\section{Introduction}
Future on-orbit servicing and active debris removal missions demand increasing levels of autonomy, particularly during rendezvous and proximity operations where a chaser must estimate the relative 6-DoF pose of a non-cooperative target. Vision-based pose estimation is attractive because it can provide rich information using compact sensors, but achieveing robust performances in orbit remains difficult. Recent work has shown great progress with deep learning-based monocular pose estimation, supported by dedicated datasets and benchmarks; however, deployment-relevant concerns such as extreme illumination, generalization across domains, and onboard computing constraints remain major obstacles \cite{Pauly2023SurveyMonocularSpacecraftPose,Park2021SPEEDplus}.

A recurring limitation of frame-based pipelines is their sensitivity to the space imaging regime: strong specularities, hard shadows, and rapid brightness transitions can degrade feature extraction and tracking, while motion blur and saturation can corrupt individual frames. These effects exacerbate the gap between synthetic training data and laboratory or on-orbit imagery, which is widely reported as a key barrier to operational adoption \cite{Pauly2023SurveyMonocularSpacecraftPose}. Motivated by these challenges, recent studies have explored event sensing as a way to improve robustness and reduce the simulation-to-real discrepancy by encoding scene changes rather than absolute appearance \cite{Jawaid2022BridgingDomainGapEvents,Rathinam2024SPADES}.

Event cameras are also being actively investigated in other space-relevant vision tasks. In space situational awareness (SSA), event-based sensing has been demonstrated for detecting and tracking resident space objects, leveraging the modality’s high temporal resolution and resilience to lighting variations \cite{Cohen2019EventBasedSSA,Afshar2020EventBasedObjectDetectionSSA,Ralph2022FIESTA}. Event-based sensing has similarly been studied for star tracking, where precise temporal information can support robust centroiding and filtering \cite{Chin2019StarTrackingEventCamera,Reed2025EBSEKF}. These results motivate event-based spacecraft pose estimation as a promising direction for proximity operations.

Crucially, spacecraft are severely constrained in size, weight, and power (SWaP), and pose-estimation pipelines must often run continuously with tight latency budgets. Neuromorphic processors provide a complementary computing paradigm: by communicating activations as sparse events and operating efficiently at low precision, they can offer favorable performance-per-watt for edge perception \cite{Davies2018Loihi,Furber2014SpiNNaker,BrainChip2025AKD1000Brief}. BrainChip’s Akida platform is particularly relevant because it supports conversion of conventional CNNs into event-driven models and has recently flown in low Earth orbit, underscoring its potential for spaceborne AI workloads \cite{BrainChip2024AkidaOrbit}. However, despite rapid progress in both event-based space perception and neuromorphic edge AI, an end-to-end demonstration of spacecraft pose estimation on Akida-class neuromorphic hardware has not been established.

This work investigates whether event-based spacecraft pose estimation can be executed efficiently on neuromorphic hardware, and what representation/model design choices best support this goal. We adopt a hybrid pose pipeline in which a compact network regresses 2D keypoints from event-derived frames, and a geometric PnP solver recovers the final 6-DoF pose.

\vspace{-1em}
\paragraph{Contributions.}
\begin{itemize}
    \item We present an end-to-end spacecraft pose-estimation pipeline that runs keypoint regression on BrainChip Akida, demonstrating the viability of event-based pose estimation on commercially available neuromorphic hardware.
    \item We develop hardware-aware keypoint regression models for Akida V1 and Akida V2 (cloud-simulated), including quantization-aware training and conversion to Akida-compatible spiking networks.
    \item We benchmark multiple event-frame representations under single pipeline to characterize the accuracy--efficiency trade-offs most relevant to spaceborne deployment.
\end{itemize}

\section{Related Work}
\subsection{Event-based vision in space applications}
Event cameras have gained attention in space applications where conventional imaging is challenged by large brightness variations and fast apparent motion. In SSA, early work established event-based sensing as a practical modality for detecting and tracking resident space objects, demonstrating robustness to illumination dynamics and the potential for efficient processing \cite{Cohen2019EventBasedSSA}. Subsequent work expanded these ideas with event-based object detection and tracking pipelines tailored to SSA settings \cite{Afshar2020EventBasedObjectDetectionSSA}, and learning-based feature consolidation/tracking methods designed for real-time event streams \cite{Ralph2022FIESTA}.

Event sensors have also been explored for star tracking, a foundational spacecraft navigation function. Chin \textit{et al.} demonstrated star tracking with an event camera, showing that asynchronous events can support robust star observations \cite{Chin2019StarTrackingEventCamera}. More recently, Reed \textit{et al.} proposed an event-based star tracking pipeline with an Extended Kalman Filter, highlighting the synergy between event measurements and state-space filtering \cite{Reed2025EBSEKF}. These lines of work motivate event-based sensing for relative navigation and pose estimation in proximity operations.

In the specific context of spacecraft pose estimation, event sensing has been proposed as a way to mitigate the domain gap between synthetic training and real imagery \cite{Jawaid2022BridgingDomainGapEvents}. SPADES further supports this direction by providing a realistic, large-scale dataset for event-based spacecraft pose estimation with both simulated and real event data \cite{Rathinam2024SPADES}. Our work builds on these efforts, focusing on deployment-oriented design choices for neuromorphic execution.

\subsection{Monocular 6-DoF spacecraft pose estimation with standard cameras}
Monocular spacecraft pose estimation has seen rapid progress driven by deep learning and new benchmarks. A recent survey by Pauly \textit{et al.}~\cite{Pauly2023SurveyMonocularSpacecraftPose} categorizes modern methods into (i) direct end-to-end regression of the pose and (ii) hybrid modular pipelines that combine learned intermediate predictions (e.g., keypoints) with geometric pose solvers. Hybrid approaches are often preferred because they leverage projective geometry constraints and can be more data-efficient and interpretable than direct regression~\cite{Pauly2023SurveyMonocularSpacecraftPose}.

A major driver of research has been the availability of datasets spanning synthetic rendering and hardware-in-the-loop validation. SPEED+ extends prior dataset efforts toward higher fidelity and improved validation pathways, supporting work that targets operational constraints beyond pure accuracy \cite{Park2021SPEEDplus}. Despite this progress, robustness under extreme illumination and generalization across domains remain key challenges for monocular pipelines \cite{Pauly2023SurveyMonocularSpacecraftPose}, motivating alternative sensing modalities such as event cameras.

\subsection{Neuromorphic hardware and edge deployment}
Neuromorphic processors aim to provide energy-efficient computation by using event-driven communication and spiking neural representations. Intel’s Loihi demonstrated a manycore digital neuromorphic design with on-chip learning primitives, enabling efficient execution of spiking workloads \cite{Davies2018Loihi}. SpiNNaker represents a complementary architecture built for scalable spike-based computation with efficient event routing \cite{Furber2014SpiNNaker}. These platforms have spurred interest in deploying spiking models for perception and control under tight power budgets.

BrainChip’s Akida platform targets edge inference by running event-domain networks with low-precision weights/activations and sparse internal activity \cite{BrainChip2025AKD1000Brief}. Within remote sensing, Lenz and McLelland demonstrated low-power ship detection from satellite imagery using Akida, illustrating the relevance of neuromorphic processing for bandwidth- and power-constrained aerospace workloads \cite{Lenz2024LowPowerShipDetection}. BrainChip has also reported the launch of Akida technology into low Earth orbit, indicating early space heritage for the platform \cite{BrainChip2024AkidaOrbit}. Our work complements these developments by focusing on spacecraft relative pose estimation---a core autonomy function for proximity operations---and demonstrating it on Akida hardware.

\section{Methodology}
\subsection{System overview}
We implement a hybrid pose-estimation pipeline that follows a common structure in spacecraft pose estimation: (i) obtain a target region-of-interest (ROI), (ii) regress 2D keypoints within the ROI, and (iii) recover the 6-DoF pose using a Perspective-$n$-Point (PnP) solver. This paper focuses on stages (ii) and (iii), with the keypoint regression network designed to run on Akida. The ROI may be produced by a separate detector (not the focus here) or taken from ground-truth for controlled benchmarking.

\subsection{Event-frame representations}
Event cameras output an asynchronous stream of events, each represented as $\mathbf{e}=(x,y,p,t)$ where $(x,y)$ are pixel coordinates, $p\in\{-1,+1\}$ is polarity, and $t$ is the timestamp. For CNN/SNN processing, we convert events in a temporal accumulation window into an image-like tensor. We focus on three lightweight representations that are compatible with compact models and hardware execution, following prior event-vision practice and the SPADES setup \cite{Gallego2022EventBasedVisionSurvey,Rathinam2024SPADES}:

\begin{itemize}
    \item \textbf{Event-to-Frame (E2F):} a polarity-coded image showing the most recent event per pixel within the window. Pixels are assigned a value/color based on the polarity of the last event (with a default background value when no event occurred).
    \item \textbf{2D Histogram (2DHist):} an event count image obtained by accumulating the number of events per pixel over the window (optionally with polarities merged or stored in separate channels), followed by normalization.
    \item \textbf{Locally-Normalised Event Surfaces (LNES):} a two-channel (positive/negative) representation that accumulates events while weighting them by normalized timestamp within the window, thereby preserving coarse temporal ordering \cite{Rathinam2024SPADES}.
\end{itemize}

These representations trade temporal fidelity against simplicity and computational cost; we evaluate them under a consistent pose-estimation pipeline.

\subsection{Keypoint regression networks}
Given an ROI event-frame tensor, the network predicts the 2D locations of a predefined set of spacecraft keypoints.

\paragraph{Akida V1 (coordinate regression).}
For Akida V1, we use a MobileNet-style architecture constrained to operations supported efficiently by the platform (standard and depthwise-separable convolutions followed by a small fully-connected head). The model directly regresses $2K$ scalars for $K$ keypoints (x/y per keypoint). This choice avoids high-resolution decoder layers typically required for dense heatmap prediction, which can be difficult to fit within V1 resource constraints.

\paragraph{Akida V2 (heatmap regression).}
To exploit Akida V2 capabilities (evaluated via Akida Cloud simulation), we also implement an encoder--decoder network that outputs $K$ heatmaps, one per keypoint. Heatmaps provide a spatial probability distribution and often improve localization robustness in keypoint pipelines, at the cost of higher compute and memory. The V2-oriented design leverages architectural blocks better supported by the newer toolchain/hardware model (e.g., skip connections and transpose convolutions, when available).

\subsection{Pose recovery via PnP}
Given predicted 2D keypoints $\{\hat{\mathbf{u}}_i\}_{i=1}^{K}$, known corresponding 3D keypoints $\{\mathbf{X}_i\}_{i=1}^{K}$ in the target body frame, and camera intrinsics, we recover the relative pose $(\mathbf{R},\mathbf{t})$ by solving a standard PnP problem. In practice, any established PnP solver (e.g., EPnP, optionally with robust estimation) can be used; this work emphasizes the accuracy/efficiency of the keypoint stage under neuromorphic constraints.

\subsection{Quantization, SNN conversion \& Deployment}
We train the networks in floating point, then apply Quantization-Aware Training (QAT) to meet Akida precision constraints. For Akida V1, we use low 4-bit weights/activations (with an optional higher 8 bit precision for the first layer input/weights when supported) and convert the resulting model to an Akida-compatible spiking network using the BrainChip toolchain \cite{BrainChip2025AKD1000Brief}. Akida V2 supports the 8-bit quantization for all network parameters, limiting accuracy decrease during deployment. We benchmark throughput and accuracy directly on Akida V1 hardware for the coordinate-regression model, and on Akida Cloud for the V2 heatmap model to characterize performance on the next-generation platform.

\section{Experiments and Results}

\subsection{Dataset and Preprocessing}
We utilize the SPADES dataset \cite{Rathinam2024SPADES}, which provides a comprehensive collection of simulated and laboratory-captured event streams for spacecraft proximity operations. The dataset contains 179,400 synthetic event frames derived from 300 unique trajectories and 15,500 real event frames from 31 trajectories captured using a Prophesee EVK4-HD camera. However, for this study, we only used the synthetic dataset, the real data is not considered for evaluation. To evaluate model performance and convergence, we partition the synthetic dataset into three subsets:
\begin{itemize}
    \item \textbf{Training Set:} 107,634 frames used for the initial floating-point training and subsequent Quantization-Aware Training (QAT).
    \item \textbf{Validation Set:} 35,877 frames utilized for hyperparameter tuning and model selection.
    \item \textbf{Test Set:} 35,889 frames reserved for final performance characterization and hardware benchmarking.
\end{itemize}

Ground truth labels consist of the relative 6-DoF pose, represented by a translation vector $\mathbf{t} \in \mathbb{R}^3$ and a unit quaternion $\mathbf{q}$. We define a set of $K=8$ keypoints corresponding to the corners of the target spacecraft bus.

For the keypoint regression network input, a Region-of-Interest (ROI) is cropped around the target spacecraft bus using ground-truth bounding boxes. These crops are resized to a fixed spatial resolution of $224 \times 224$ pixels. Events are accumulated into frames using a fixed temporal window of $\Delta t = 50\,\text{ms}$, resulting in input tensors that are compatible with the convolutional backbones of the Akida V1 and V2 platforms. The dataset specifically includes challenging space-representative illumination conditions, such as high-intensity specularities and hard shadows, to test the robustness of the event-based modality.

\subsection{Evaluation metrics}
We evaluate the pipeline using standard spacecraft pose metrics. Let $(\tilde{\mathbf{t}}, \tilde{\mathbf{q}})$ and $(\mathbf{t},\mathbf{q})$ represent estimated and ground-truth translation and unit quaternions, respectively.
\begin{itemize}[itemsep = 0.15em]
    \item \textbf{Translation error ($E_T$):} $\lVert \tilde{\mathbf{t}} - \mathbf{t} \rVert_2$. We also report the normalized error $E_T^{norm} = E_T / \lVert \mathbf{t} \rVert$.
    \item \textbf{Rotation error ($E_R$):} $2\arccos\bigl|\langle \tilde{\mathbf{q}}, \mathbf{q} \rangle\bigr|$.
    \item \textbf{Pose error ($E_P$):} $E_{P} = E_R[\text{rad}] + E_T^{norm}$, as in \cite{Park2023SPEC2021}.
    \item \textbf{Percentage of correct keypoints ($PCK$):} The fraction of predicted keypoints within a $d$-pixel threshold of the ground truth.
    \item \textbf{Pose rejection ($Rej$):} To maintain the integrity of the performance characterization, we define a rejection criterion based on solver convergence and physical plausibility. An estimated pose is discarded if the PnP solver fails to yield a numerical solution or if the resulting translation magnitude exceeds a safety threshold of $\lVert \tilde{\mathbf{t}} \rVert > 30\,\text{m}$. These rejected samples are recorded separately and are excluded from aggregate error calculations ($E_T$, $E_R$, and $E_P$).
\end{itemize}

\subsection{Architectural baselines and training}

\subsubsection{Akida V1 coordinate regression}
For the Akida V1 baseline, we adopted a direct coordinate regression approach using a modified MobileNet backbone consisting of 1.88\,M parameters ($\sim$7.19\,MB). The network processes event-frame representations through 12 separable convolutional blocks. A global flatten layer followed by a dense head regresses 16 scalar values, corresponding to the Cartesian $(x, y)$ coordinates of the 8 keypoints. Models were trained for 300 epochs using the synthetic SPADES dataset (107,634 samples) with a multi-GPU MirroredStrategy to ensure convergence of the scalar outputs. All 8 keypoints were used for the PnP optimization stage.

\subsubsection{Akida V2 heatmap regression}
To exploit the enhanced capabilities of the Akida V2 platform, we implement a hardware-aware encoder--decoder architecture optimized for spatial heatmap regression. The network utilizes a MobileNetV2-style backbone and a decoder sub-network employing transposed convolutions to regress $56 \times 56 \times 8$ heatmaps. This model consists of 1,718,410 parameters ($\sim$6.56\,MB). 

By outputting spatial probability distributions rather than absolute coordinates, the network achieves significantly faster convergence; validation PCK scores $>0.96$ were reached in only 60 epochs (one-fifth of the training duration required for V1). To ensure fidelity on neuromorphic hardware, the network was trained using Quantization-Aware Training (QAT) to 8-bit precision. With the heatmaps allows selection of confident keypoints, we performed PnP optimization with both top-5 and 8 (all) keypoints.

\subsection{Pose estimation performance}
Tables \ref{tab:v1results} and \ref{tab:v2results} summarize the performance of the Akida V1 and Akida V2 models.

\paragraph{V1 analysis (coordinate regression).} 
Floating-point V1 models achieve competitive SPEED scores ($\sim$0.036), but performance degrades significantly upon 4-bit quantization. The PCK drops from $>93\%$ to $\sim$30\% for E2F and 2DHist, while the LNES representation collapses entirely (PCK 0.20, 2370 rejections). This indicates that direct scalar regression is highly sensitive to precision loss in the fully connected Akida V1 layers.

\paragraph{V2 analysis (heatmap regression).} 
The Akida V2 models show remarkable robustness to quantization. The transition from floating-point to quantized weights results in negligible accuracy loss; the quantized E2F model achieves a SPEED score of 0.0208, even slightly outperforming its floating-point counterpart. This suggests that heatmap regression is inherently more compatible with low-precision neuromorphic activations.

\subsection{Impact of event representations}
\begin{figure*}[t]
    \centering
    \begin{subfigure}{0.48\textwidth}
        \centering
        \includegraphics[width=\linewidth]{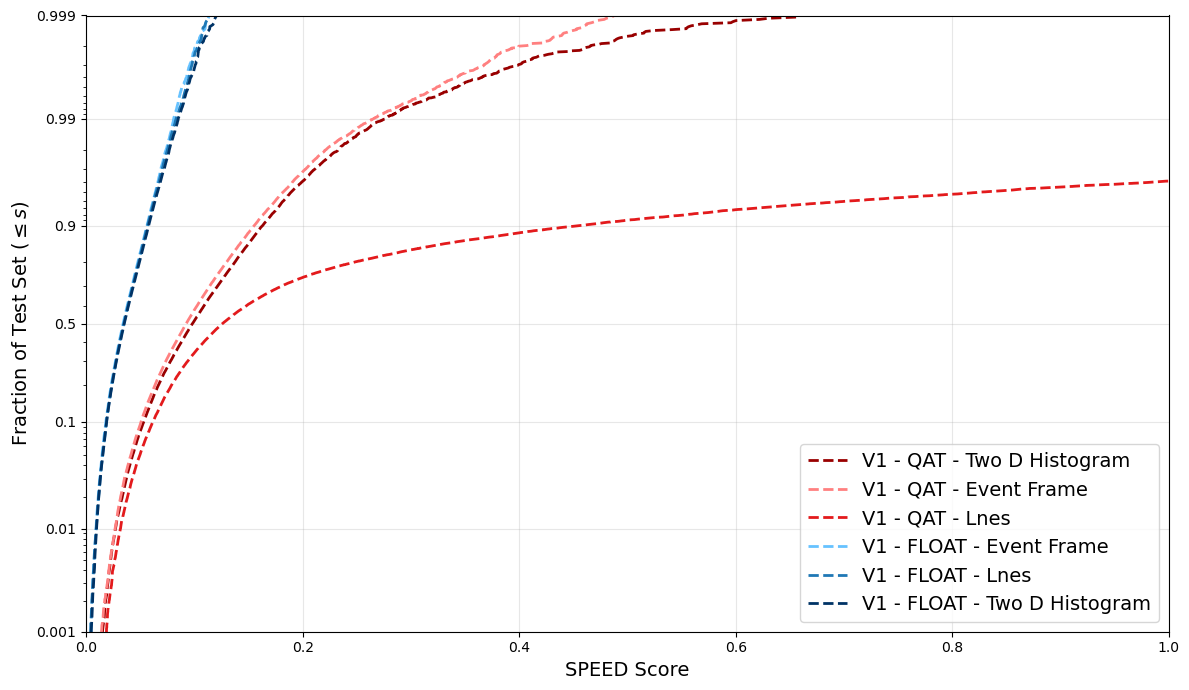}
        \caption{Akida V1: Float vs. Quantized}
        \label{fig:akida_v1_compare}
    \end{subfigure}
    \hfill 
    \begin{subfigure}{0.48\textwidth}
        \centering
        \includegraphics[width=\linewidth]{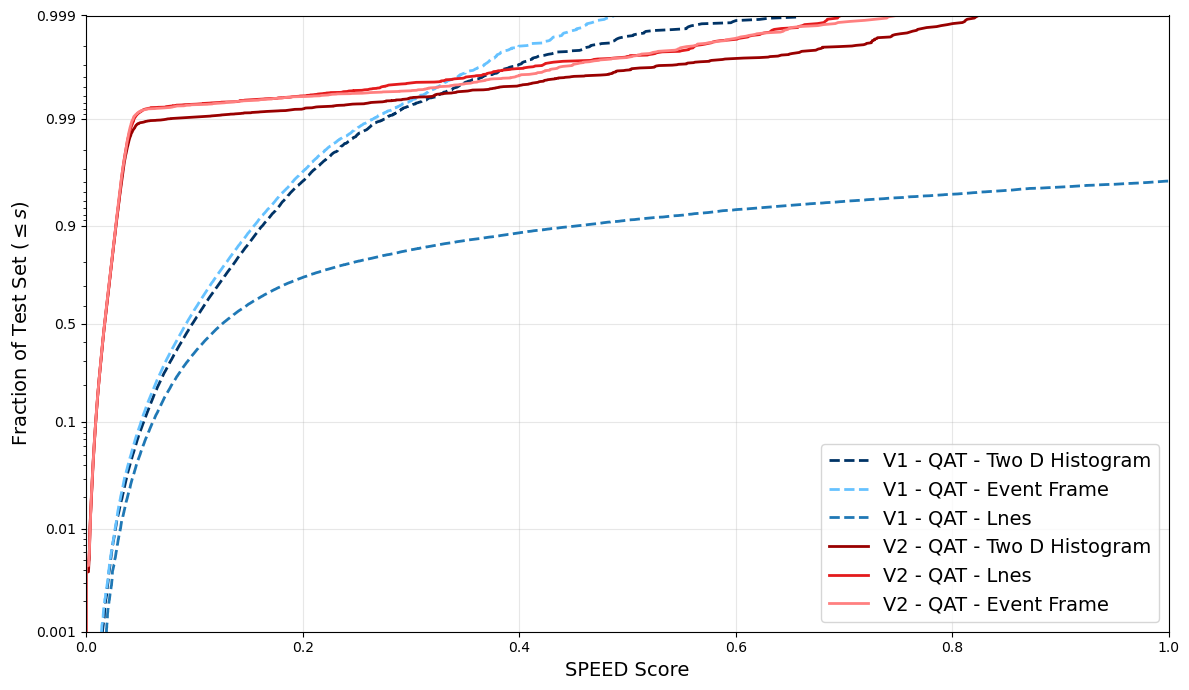}
        \caption{Akida V1 vs. V2: Quantized Comparison}
        \label{fig:akida_v1_vs_v2_quantized}
    \end{subfigure}

    \caption{Cumulative distribution of SPEED scores on the SPADES dataset. (a) shows the impact of quantization on various representations for Akida V1, while (b) compares the performance of quantized Akida V1 and V2 architectures.}
    \label{fig:combined_speed_scores}
\end{figure*}

Figures \ref{fig:akida_v1_compare} and \ref{fig:akida_v1_vs_v2_quantized} illustrate the cumulative distribution of SPEED scores. While floating-point models are largely invariant to representation choice, \textbf{E2F} and \textbf{2DHist} are significantly more robust than \textbf{LNES} for V1 hardware. However, for V2 models, LNES provides the highest accuracy after quantization ($E_P = 0.0207$), suggesting that the LNES representation holds valuable temporal information that the higher-capacity V2 architecture can effectively exploit.

\subsection{Impact of Quantization}
\begin{figure*}[!h] 
    \centering
    \begin{subfigure}{\textwidth}
        \centering
        \includegraphics[width=\linewidth]{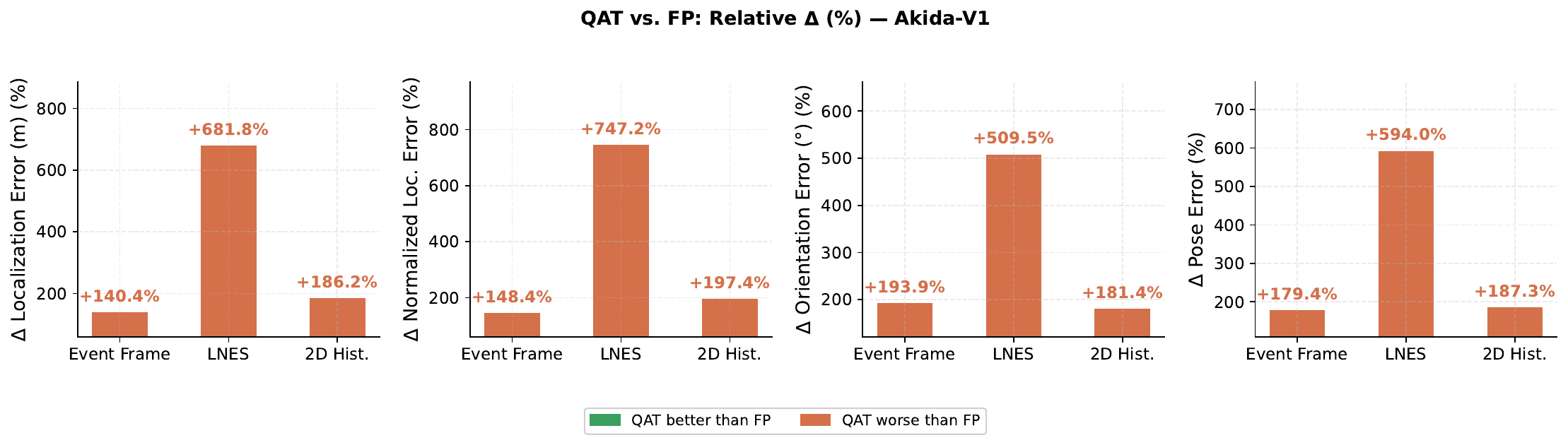}
        \label{fig:delta_v1}
    \end{subfigure}

    \vspace{-3em} 

    \begin{subfigure}{\textwidth}
        \centering
        \includegraphics[width=\linewidth]{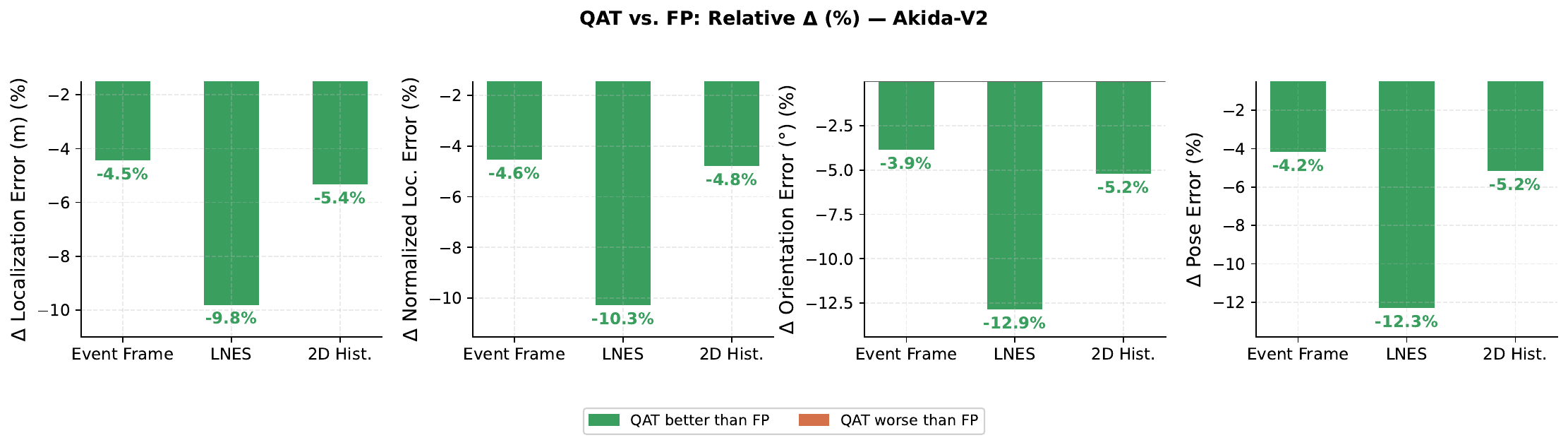}
    \end{subfigure}
         \begin{subfigure}{\textwidth}
        \centering
        \includegraphics[width=\linewidth]{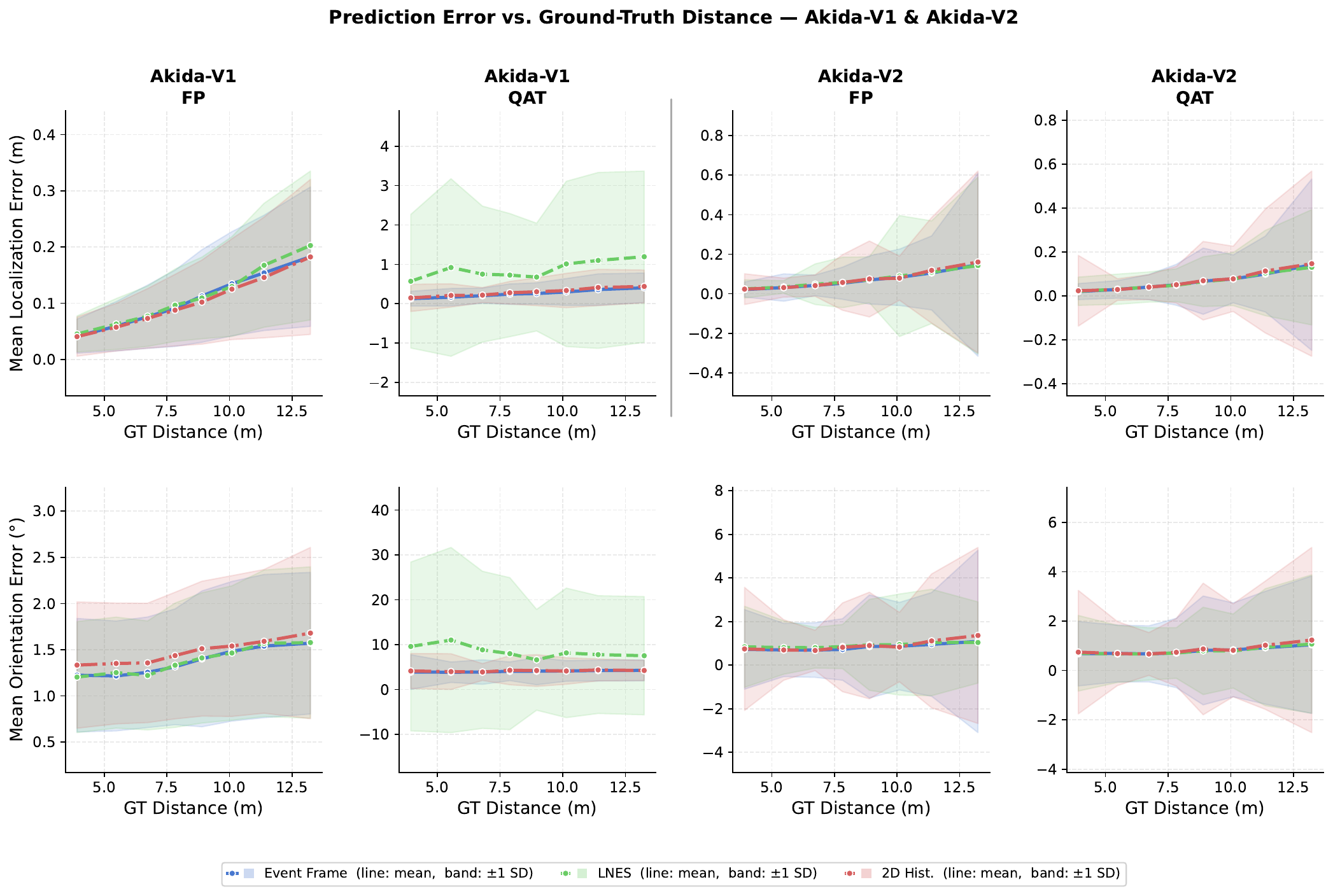}
    \end{subfigure}
\vspace{-2em}
    \caption{\textbf{Comprehensive Quantization and Error Analysis:} Top two rows display the performance impact of quantization across different representations. Bottom two rows Mean localization and orientation error binned by ground-truth object distance for full-precision (FP) and quantization-aware trained (QAT) models between V1 and V2 architectures. Shaded regions denote $\pm$1 standard deviation.}
    \label{fig:full_four_row_analysis}
\end{figure*}

\begin{figure*}[!t] 
    \centering
       \begin{subfigure}{\textwidth}
        \centering
        \includegraphics[width=\linewidth]{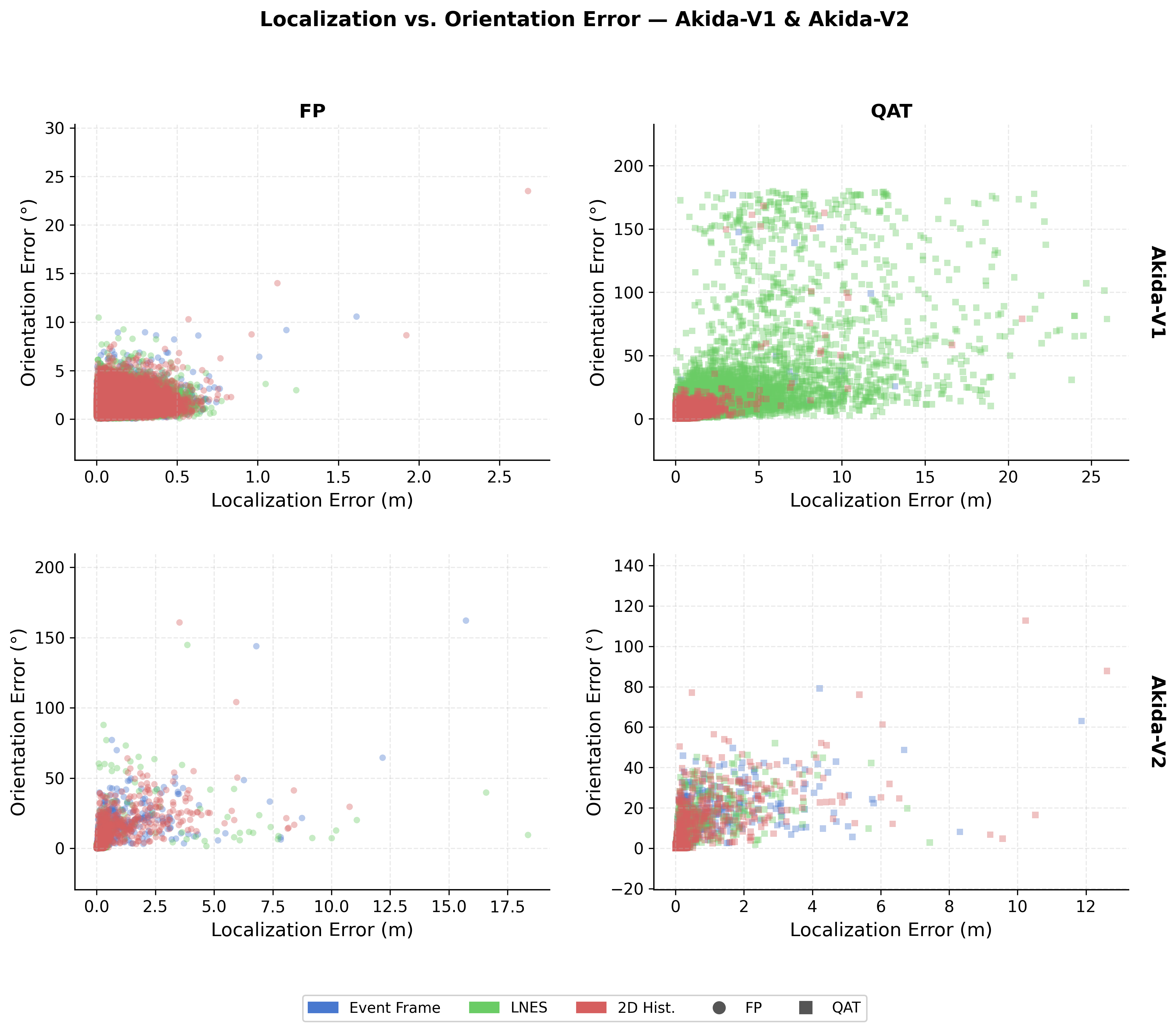}
    \end{subfigure}
    \vspace{-2em}
    \caption{Scatter distribution of per-sample localization error versus orientation error for full-precision (FP) and quantization-aware trained (QAT) models across Akida-V1 and Akida-V2 architectures. Each point represents one test sample, coloured by input representation.}
    \label{fig:combined_error_dist}
\end{figure*}

Figures \ref{fig:full_four_row_analysis} and \ref{fig:combined_error_dist} compare the performance of floating point models vs quantization models. V1's regression head turns quantization noise into direct coordinate error with no buffering, while V2's heatmap head absorbs that noise spatially and was architecturally designed and trained to be quantization-aware from the outset. Quantization precision also affects the model performance: Akida-V1 operates at 4-bit precision, while Akida-V2 supports 8-bit inference. This difference alone has a substantial impact on representational capacity and is clearly visible on the performance between both models.

\subsection{Qualitative analysis}
\begin{figure}[h]
    \centering
    \includegraphics[width=0.9\linewidth]{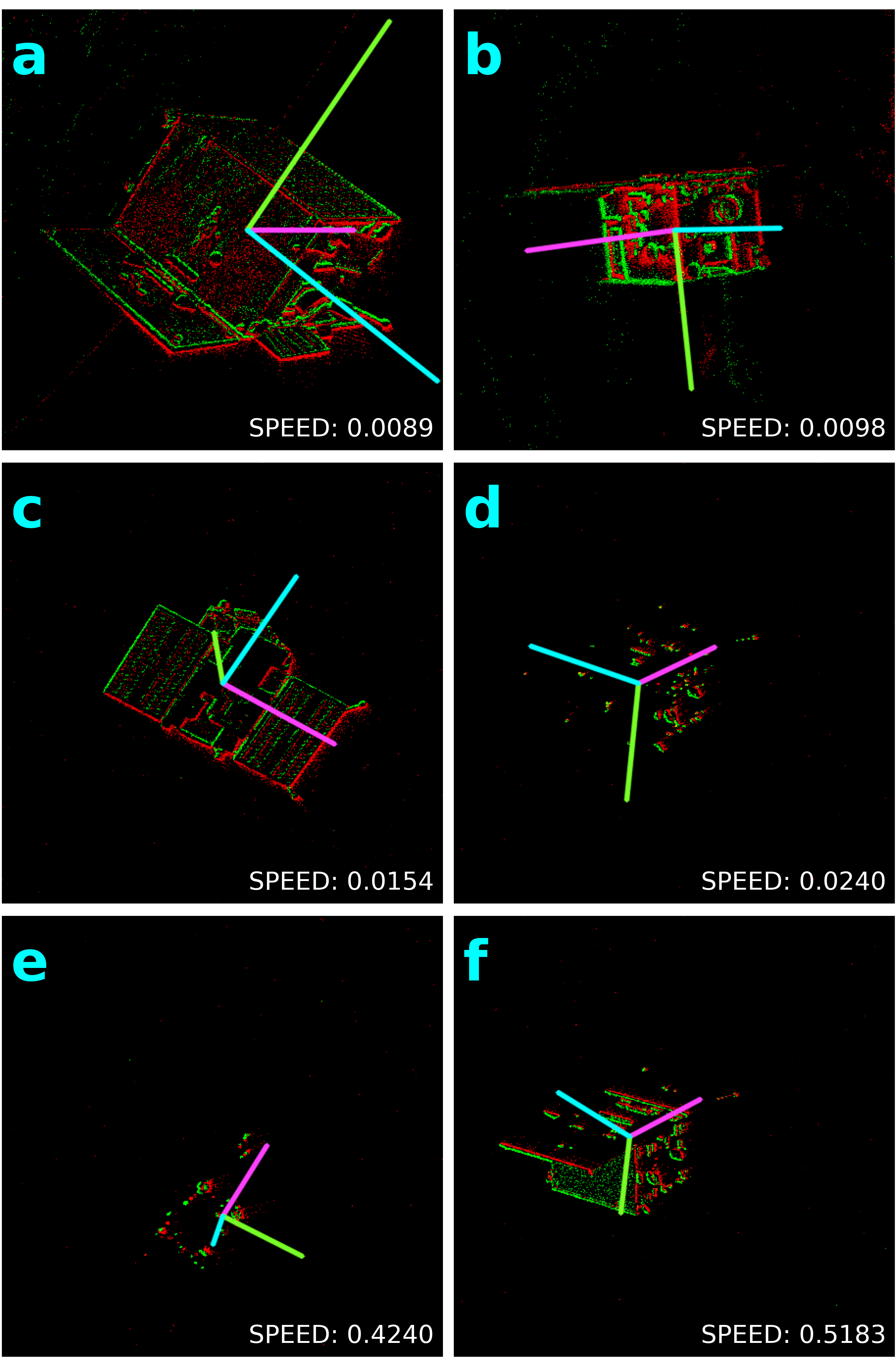}
    \caption{Qualitative results on LNES. Rows show samples with high (a-b), median (c-d), and low (e-f) SPEED scores.}
    \label{fig:lnes-grid}
    \vspace{-1em}
\end{figure}

Qualitative results (Fig. \ref{fig:lnes-grid}) demonstrate that pose errors are driven primarily by illumination. High-performing samples typically feature well-defined edges on the satellite bus, whereas poor samples coincide with high-noise regions or sparse event clusters at high ranges. Variations in lighting conditions remain the most significant factor that impacts model performance.

\begin{table}[t]
\centering
\caption{Comparison of pose estimation performance of AKIDA V1 models across different event representation. }
\vspace{-0.2cm}

\resizebox{\columnwidth}{!}{%
\begin{tabular}{c | c c c c c | c }
\toprule
 \textbf{Event}  & \textbf{Keypoint} & \multicolumn{4}{c}{\textbf{Pose Metrics}} & \textbf{Rej}\\
   \textbf{Rep.}     & $PCK  \uparrow$ & $E_{T} [m] \downarrow$ &  ${E}_{T}^{norm} [.] \downarrow$ & $E_{R}$ [\textdegree ] $\downarrow$ & $E_{pose} [.] \downarrow$  & \\
              \midrule
    \multicolumn{7}{c}{\textbf{V1 Float Models}} \\
      LNES & 0.9424 & 0.1114  &  0.0128 &   1.3793 &  0.0367 & 1 \\
      E2F & 0.9439 & 0.1065 &  0.0122 &  1.3759  &   0.0361 & 1 \\
      2DHist & 0.9303 & 0.1018 &  0.0117 &  1.4752 &   0.0373 & 0\\
          \midrule
    \multicolumn{7}{c}{\textbf{Quantized}} \\
      LNES &  0.2036 &0.8710 & 0.1083 & 8.4068 &  0.2550 & 2370 \\
      E2F & 0.3328  & 0.2560 & 0.0304  &  4.0433  &  0.1009 & 162 \\
      2DHist & 0.3094 & 0.2913 &  0.0348  &  4.1512 &   0.1072 & 165 \\
          \midrule
             \multicolumn{7}{c}{\textbf{Deployed}} \\
      LNES & 0.2110 & 0.8468 & 0.1072  & 8.7214  & 0.2594  & 259 \\
      E2F & 0.3344 & 0.2617 & 0.0316  &  4.0106 &  0.1016 & 17\\
      2DHist & 0.3127 & 0.3063 & 0.0363  & 4.1123  & 0.1081  & 13\\
         \bottomrule   
\end{tabular}
}
\label{tab:v1results}

\end{table}

\begin{table}[t]
\centering
\caption{Comparison of pose estimation performance of AKIDA V2 models across different event representation}
\vspace{-0.2cm}

\resizebox{\columnwidth}{!}{%
\begin{tabular}{c | c c c c c | c }
\toprule
 \textbf{Event}  & \textbf{Keypoint} & \multicolumn{4}{c}{\textbf{Pose Metrics}} & \textbf{Rej}\\
   \textbf{Rep.}     & $PCK  \uparrow$ & $E_{T} [m] \downarrow$ &  ${E}_{T}^{norm} [.] \downarrow$ & $E_{R}$ [\textdegree ] $\downarrow$ & $E_{pose} [.] \downarrow$  &\\
              \midrule
    \multicolumn{7}{c}{\textbf{V2 Float Models}} \\
      LNES (5) & 0.9669 & 0.0719 &   0.0080 & 0.9084  &  0.0236 & 0 \\
    LNES (8) & 0.9669 &  0.0704 & 0.0078  & 0.8301  & 0.0220  & 3 \\
      E2F (5) & 0.9684 &  0.0722  &  0.0080 &  0.9252 &   0.0239 &  0 \\
      E2F (8) & 0.9684  &  0.0690 &  0.0076  & 0.8261  & 0.0217  & 3 \\
      2DHist (5) & 0.9668 & 0.0682 & 0.0077  &   0.9274 &   0.0236 & 0 \\
    2DHist (8) & 0.9668 &  0.0730 &  0.0080 & 0.8932  &  0.0233 & 13 \\
         \bottomrule   
          \midrule
    \multicolumn{7}{c}{\textbf{Quantized}} \\
      LNES (8) & 0.9704 &  0.0648 &  0.0072 & 0.7914  &  0.0207 & 5 \\
      E2F (8) &  0.9712 & 0.0660 & 0.0072  &  0.7941  &  0.0208  & 6 \\
      2DHist (8) & 0.9688 &  0.0691 &  0.0076  &  0.8464 &   0.0221 & 9\\
         \bottomrule   
          \midrule
    \multicolumn{7}{c}{\textbf{Deployed - AKIDA cloud [1000 samples]}} \\
      LNES (8) & - & 0.0714 & 0.0076 & 0.9393  &  0.0237  & 0 \\
      E2F (8) & - &  0.0588 & 0.0065  &  0.7171  & 0.0188  & 0 \\
      2DHist (8) & - & 0.0747  &  0.0082 & 0.9723   &  0.0249  & 0 \\
         \bottomrule  
\end{tabular}
}
\label{tab:v2results}

\end{table}

\begin{table}[!h]
\centering
\caption{Comparison of power and latency metrics for Akida V1 and Jetson Orin Nano. Tests are run for 4000 samples per representation and average values are given.}
\label{tab:power_comparison}
\vspace{-0.2cm}
\resizebox{0.8\columnwidth}{!}{%
\begin{tabular}{l c c c}
\toprule
{Event} & {Dynamic Power} & {Idle Power} & {Speed} \\
{Rep.}  & {[mW] $\downarrow$} & {[mW] $\downarrow$} & {[fps] $\uparrow$} \\
\midrule
\multicolumn{4}{c}{\textit{Jetson Orin Nano (Floating Point)}} \\
\midrule
LNES   & 2491 & 3997 & 321 \\
E2F    & 2519 & 3983 & 330 \\
2DHist & 2650 & 4009 & \textbf{349} \\
\midrule
\multicolumn{4}{c}{\textit{Akida V1 (Deployed Quantized)}} \\
\midrule
LNES   &  \textbf{917} & \textbf{927} & 16 \\
E2F    &  919 & \textbf{927}  & 16  \\
2DHist & 928 & \textbf{927} & 17  \\
\bottomrule
\end{tabular}
}
\vspace{-1em}
\end{table}

\subsection{Hardware benchmarking}
We benchmark the throughput and power on the AKD1000 (V1) SoC and the Akida Cloud simulation (V2). Our results are compared with a NVIDIA Jetson Orin Nano board, a common platform for edge uses cases, in \ref{tab:power_comparison}. The Jetson Nano board runs the floating point model converted to a tensorrt engine for optimized speed and power performances. As expected, both static and dynamic power is 3 to 4 times lower using the AKD1000 chip. However, inference speed is much lower than on the Jetson board (17 forward passes per second while the jetson 349). One explanation of this difference is that Akida chip interface with the host computer has not been optimized yet and doest not allow to fully leverage its fast, even based processing capabilities. 

\paragraph{Latency and Complexity.}
The Akida V1 model achieves \textbf{17 FPS} on the physical AKD1000. In contrast, the Akida V2 heatmap model achieves \textbf{2.43 FPS} (25MHz). This discrepancy is explained by architectural complexity: the V2 model utilizes an encoder--decoder structure with transposed convolutions, whereas the V1 regressor is a compact backbone. However, as shown in Table~\ref{tab:hardware_stats}, assuming linear frequency scaling, the V2 model is projected to reach 37.21 FPS at 400 MHz. This estimate does not account for potential bandwidth or routing constraints and represents an ideal scaling scenario.

\paragraph{Sparsity-Power Trade-off.}
On the AKD1000, power consumption is strongly correlated with input activation sparsity. The \textbf{2DHist} and \textbf{E2F} representations resulted in lower dynamic power draw compared to \textbf{LNES} due to their sparser activation patterns. Power consumption on the Akida platform is inherently event-driven; with the observed cycle counts and high activation sparsity, the estimated power draw remains within the milliwatt range typical for edge neuromorphic processors.

\begin{table}[!h]
\centering
\caption{Akida V2 Hardware Metrics: Measured (25\,MHz) and Projected (400\,MHz)}
\label{tab:hardware_stats}
\resizebox{\columnwidth}{!}{
\begin{tabular}{l c c c c}
\toprule
\textbf{Representation} & \textbf{Latency} & \multicolumn{2}{c}{\textbf{Throughput (FPS) $\uparrow$}} & \textbf{Params} \\
\cmidrule(lr){3-4}
 & \textbf{($10^6$ Cycles)} & \textbf{@ 25 MHz} & \textbf{@ 400 MHz} & \\
\midrule
Event Frame (EF) & 11.04 & 2.18 & 36.78 & 1.71M \\
2D Histogram    & 10.70 & 2.33 & 37.21 & 1.71M \\
LNES            & 11.03 & 2.25 & 36.04 & 1.71M \\
\bottomrule
\end{tabular}%
}
\end{table}

\vspace{-1em}
\section{Discussion}
The transition from floating-point training to quantized deployment reveals a significant performance gap tied to regression methodology. For Akida V1, hardware limitations forced the use of direct keypoint regression, which proved highly vulnerable to quantization noise. In contrast, Akida V2 enabled heatmap regression, which transforms the problem into a spatial classification task. This higher-capacity architecture proved far more resilient to low-precision activations. Furthermore, the event-driven nature of the hardware allows for a "power-on-demand" approach where sparser representations like E2F reduce the thermal footprint of the processor.

\section{Conclusion}
This work demonstrated the first end-to-end event-based spacecraft pose-estimation pipeline on BrainChip Akida hardware. While quantization remains a hurdle for direct coordinate-regression, encoder--decoder heatmap networks on Akida V2 provide a robust pathway for autonomous navigation. Future work will investigate recurrent architectures to further leverage the temporal dynamics of event streams and evaluate the pipeline on physical Akida V2 as it becomes available.

\section*{Acknowledgement}

This research was partially funded by the Luxembourg National Research Fund (FNR), project reference C21/IS/15965298/ELITE. The work was started during the NeuroTUM Hackathon'24 in Munich, hosted by Fortiss GmbH, where Roua Brini, Jakub Kuspien and Manoj Bhat made valuable contributions. The authors thank Gilles Bézard (BrainChip), Alf Kuchenbuch (BrainChip), and Priyadarshini Kannan (Fortiss) for their assistance with the Akida BrainChip platform.

{
    \small
    \bibliographystyle{ieeenat_fullname}
    \bibliography{library}
}


\end{document}